\documentclass{article}

\usepackage{lscape}
\usepackage{arxiv}
\usepackage{longtable}
\usepackage[utf8]{inputenc} 
\usepackage[T1]{fontenc}    
\usepackage{hyperref}       
\usepackage{url}            
\usepackage{booktabs}       
\usepackage{amsfonts}       
\usepackage{nicefrac}       
\usepackage{microtype}      
\usepackage{lipsum}
\usepackage{graphicx}
\graphicspath{ {./images/} }

\title{Impact analysis of recovery cases due to
COVID-19 outbreak using deep learning
model}

\author{
 Md Ershadul Haque \\
 School of Computing, Mathematics and Engineering\\
  Charles Sturt University \\
 Bathurst, Australia \\
  \texttt{mhaque@csu.edu.auu} \\
   \And
 Samiul Hoque \\
  Department of Electrical and Electronic Engineering\\
  Feni University\\
  Feni, Bangladesh \\
  \texttt{samiul0906@gmail.com} \\
  \And
}

\begin{document}
\maketitle
\begin{abstract}
The present world is badly affected by novel coronavirus (COVID-19). Using medical kits to identify the coronavirus affected persons are very slow. What happens in the next, nobody knows. The world is facing erratic problem and don't know what will happen in near future. This paper is trying to make prognosis of the coronavirus recovery cases using LSTM(Long Short Term Memory). This work exploited data of 258 regions, their latitude and longitude and the number of death of 403 days ranging from 22-01-2020 to 27-02-2021. Specifically, advanced deep learning-based algorithms known as the LSTM, play a great effect on extracting highly essential features for time series data (TSD) analysis.There are lots of methods which already use to analyze propagation prediction. The main task of this paper culminates in analyzing the spreading of Coronavirus across worldwide recovery cases using LSTM deep learning-based architectures. 
\end{abstract}


\section{Introduction}\label{sec1}

Now a days, COVID-19 is a global pandemic to the health issue word-wide. Not only, to combat to the unknown disease is a greatest challenge.  Broadly speaking, human are affected by COVID-19. This virus belongs to Nidovirales which consist of Arteriviridae, Roniviridae, and Coronaviridae \cite{c24, c25}. In the meantime, the family of the Coronaviridae  is divided into Coronavirinae and Torovirinae. Furthermore, it is categorized into four types of variant's such as gamma, delta, alpha and beta. It is one kind of RNA genome virus. The dimension of this virus is 26 to 32 kilobases and visible  by the electron microscope. World Health Organization (WHO) conducted several research on COVID-19 to predict the new strain of COV named as COVID-19. The peoples are getting infected more and the illness is spreading tremendously word-wide. Therefore, the word is facing threat due to the pandemic of COVID-19 \cite{c26} 
     
COVID-19 was started in Wuhan city, capital of Hubei provenance in China at the end of 2019 December while Chinese peoples are getting preparation to celebrate their spring festival\cite{c1}. For the outbreak of Covid 19 later it was named as novel coronavirus \emph{(Covid-19)} by the WHO(Word Health Organization). Hundred of thousand infected people left Wuhan which is the central hub of the provenance and they are identified as first transmission media  to carry this virus. The virus SARS-CoV2, causing the Coronavirus disease (COVID-19), has spread from Wuhan, China to every populated continent\cite{c2}. As the COVID-19 outbreak is now a pandemic, it will be important to have tools to rapidly identify those at risk of morbidity and mortality.  We must augment clinical skills to identify mild cases that will progress to critical illness\cite{c3}. 

Predictive analysis (a form of artificial intelligence) learns from historical data to help forecasting future outcomes. This analysis uses machine learning algorithms that can extract insights and rules from recorded data with the most predictive tool for making accurate predictions. The analysis results in predicting the tentative future contexts that can be conducive. At present, its the demands of time what will be the scenario of COVID-19 in future. Everyday people's are getting sick due to pandemic of COVID-19 and word is facing a public health threat\cite{c26}. In this situation, the prevent of COVID-19 is an emergency needed. By the way, still now there is no exact cure. In such situation, machine learning approach is a viable solution to find the exact battle to fight against the COVID-19. Moreover, the impact of recovery cases  is a vital issue. Many works have already been done using machine learning. In \cite{shah2020haar} dynamics of susceptible, infected and recovered cases are studied. All the cases are simulated considering both on and off immigration condition. All the plots here are shown from mathematical model. So there may be discrepancies from the practical data. The work in \cite{khan2022optimal} Covid-19 outbreak has been modeled mathematically by introducing Caputo Fractional derivatives. In \cite{khan2022numerical} a compartmental model was made to formulate the epidemic. The effect of vaccination was also deemed. 

The main motivation of this work come from the analysis of the death across the world using AI based technology. The objective of this work is to predict the impact of recovery cases using LSTM deep learning model. The job done here are:
\begin{enumerate}
    \item Development of a python code by Pytorch
    \item Collection of data as CSV file from authentic source from internet.
    \item Using the code and data the prognosis of recovery number in future days was performed.
\end{enumerate}

\section{Related Work}

The virus \emph{SARS-CoV2}, which causes coronavirus disease \emph{(COVID-19)} has scattered as a pandemic form and has spread to every continent. We must augment clinical skills to identify mild cases that will progress to critical illness. The study in\cite{c4} showed that the outbreak seen has not matched the clinical severity seen in the initial Wuhan, Hubei epicenter. A clear limitation of this study is the size of the data set though the models were 70\% - 80\% predictive. 

In \cite{c5}, by the utilization of COVID-19 daily data of mainland China excluding Hubei province, comprised of the cumulative confirmed cases, the cumulative deaths, newly confirmed cases and the cumulative recovered cases. 

The aim of \cite{c6} was to assess the machine and deep learning-based architectures performance that is proposed in the present year for classification of Coronavirus images such as, X-Ray and computed tomography.  
The target of the study in \cite{c7} was to estimate the basic reproduction number of the Wuhan novel coronavirus using SEIR compartment model to characterize the early spreading of 2019-nCoV.  
In \cite{c8}, Logistic model, Bertalanffy model and Gompertz model, which are relatively simple but support the statistical law of epidemiology, are selected to predict the epidemic situation of COVID-19. 

In \cite{c9}, the analysis of the cumulative confirmed cases, cumulative deaths, and cumulative cured cases was conducted using an Elman neural network, long short-term memory (LSTM), and support vector machine (SVM).

In the work of \cite{c10}, the COVID-19 outbreak in India has been predicted based on the pattern of China using a machine learning approach. This outbreak is predicted between the third and fourth weeks of April 2020 in India to be controlled around the end of May 2020.

Taking into account the uncertainty due to the incomplete data of infective population, they applied the well-known SEIR compartmental model for the prediction was found in \cite{c11}. 

In the study of \cite{c12} COVID-19 daily new cases and deaths in the USA from two population-based data-sets were extracted. 
Study of \cite{c13} demonstrates that real-time, publicly available ensemble forecasts issued in April-July 2020 provided robust short-term predictions of reported COVID-19 deaths in the United States. 
The paper of \cite{c14} introduces an objective approach to predicting the continuation of the COVID-19 using a simple, but powerful method to do so. 

The propose of \cite{c15} was a shallow long short-term memory (LSTM) based neural network to predict the risk category of a country. A Bayesian optimization framework have been used to optimize and automatically design country-specific networks. In COVID-19 detection, a modification version of U-net network was applied to identify the covid and non-Covid case classification purpose. In this method, X-ray flim image data was considered~\cite{c16}. On the other hand, CT image based COVID-19 classification was done in \cite{c17}. VGG16, Res-Net, CNN, auto encoder neural network networks were applied to compare the computational accuracy. The spreading case of COVID-19 was analyzed in \cite{c21} for high rise countries. Mutimodal based emotional model was proposed to analyze COVID19 by Author Gupta in \cite{c22}. A threshold condition was studied in \cite{c23} for COVID19 disease free toward the global sustainability. 

Alok Negi et. al. worked on face mask detection in public places \cite{negi2021deep}\cite{negi2020face}. Malaria infection detection was done in \cite{alok2021deep}. Multi-view video summarizing was done in \cite{kumar2017f}. In \cite{ansari2018dcr} Hidden Markov Model(HMM) was utilized to detect the depression of mind. The thoracic surgery can be assisted with AI which was suggested in \cite{darbari2021requirement}. In \cite{negi2021deep1}\cite{negi2021classification} plant disease was simulated by AI. Face mask from video streaming was performed in \cite{negi2021face1}.A human action recognition  system was designed by RESNET50 that can determine the human daily activities\cite{negi2021predictive}. In \cite{kumari2019prediction} liver disease detection was proposed to reduce the instigation of death. Unlike the SVM(Support Vector Machine), ANN(Artificial Neural Network) technique, the CNN was applied for classify the cancer cell into Benign and Malignant in \cite{dabral2019cancer}.

Qingzhen Xu\cite{xu2013novel} showed that financial prediction can be done using machine learning. In \cite{xu2019novel} the saliency map of a complex image is proposed. Osteoporosis is predicted with transfer learning in \cite{xu2020fall}. Trend of stock is predicted in \cite{xu2014novel}. Drosophila embryonic gene expression computation requires accuracy and rapidity that is presented in \cite{xu2019multi}.

\begin{landscape}
\begin{longtable}[c]{|c|c|c|}
\caption{Related Work In a Nutshell}
\label{tab:my-table}\\
\hline
\multicolumn{1}{|l|}{\textbf{Reference}} &
  \multicolumn{1}{l|}{\textbf{Advantage}} &
  \multicolumn{1}{l|}{\textbf{Limitation}} \\ \hline
\endfirsthead
\multicolumn{3}{c}%
{{\bfseries Table \thetable\ continued from previous page}} \\
\hline
\multicolumn{1}{|l|}{\textbf{Reference}} &
  \multicolumn{1}{l|}{\textbf{Advantage}} &
  \multicolumn{1}{l|}{\textbf{Limitation}} \\ \hline
\endhead
Jiang et al. {[}7{]} &
  \begin{tabular}[c]{@{}c@{}}Its identify the combinations \\ of clinical characteristics of\\ COVID-19 that predict outcomes. \\ Besides, a tool with AI capabilities\\ that will predict patients at risk for\\ more severe illness on initial\\ presentation has developed.\end{tabular} &
  \begin{tabular}[c]{@{}c@{}}Limited amount of data was used. \\ Also, no use of update technology \\ AI tool such as LSTM, \\ Efficient-Net and others.\end{tabular} \\ \hline
Azimi et al.{[}8{]} &
  \begin{tabular}[c]{@{}c@{}}Estimate the reproduction of \\ diseases affected. Exponential\\ growth rate, maximum likelihood\\ was evaluated.\end{tabular} &
  \begin{tabular}[c]{@{}c@{}}The estimation is based on\\ study cases not on real case data.\\ In addition, simulated daily cases\\ are used lead the missing of real \\ result due to sampling information\\ of PCR.\end{tabular} \\ \hline
\begin{tabular}[c]{@{}c@{}}Salehi et al.\\  {[}9{]}\end{tabular} &
  \begin{tabular}[c]{@{}c@{}}Its provides art of state for\\ future of AI technology. \\ Also, X-ray and CT scan \\ data set were discussed.\end{tabular} &
  \begin{tabular}[c]{@{}c@{}}Draw only low amount of data.\\ Discuss only old model of\\  AI technique.\end{tabular} \\ \hline
\begin{tabular}[c]{@{}c@{}}Zhou et al. \\ {[}10{]}\end{tabular} &
  \begin{tabular}[c]{@{}c@{}}Estimate the basic reproduction \\ of the confirm and suspected cases. \\ Data acquire from severe acute\\  respiratory syndrome.\end{tabular} &
  \begin{tabular}[c]{@{}c@{}}Intervention was not for\\ propagation analysis. \\ Timely and effective \\ control measures was absent for \\ further transmission analysis.\end{tabular} \\ \hline
Hao et al. {[}11{]} &
  \begin{tabular}[c]{@{}c@{}}Various kinds of AI based\\ network are used to predict \\ new cured cases, confirmed \\ cases and new deaths cases.\end{tabular} &
  \begin{tabular}[c]{@{}c@{}}The proposed model is  less\\ robust since the average value \\ is larger than actual. Further\\ performance improvement is \\ required\end{tabular} \\ \hline
\begin{tabular}[c]{@{}c@{}}Tiwari et. al.\\  {[}12{]}\end{tabular} &
  \begin{tabular}[c]{@{}c@{}}Time series forecasting method\\ to forecast the death case, \\ confirm  case and recovery case.\end{tabular} &
  \begin{tabular}[c]{@{}c@{}}Unclear about which time series\\  method is used for forecasting. \\ Also, did not consider \\ socioeconomic  consideration.\end{tabular} \\ \hline
\begin{tabular}[c]{@{}c@{}}Samuel  et al.\\  {[}13{]}\end{tabular} &
  \begin{tabular}[c]{@{}c@{}}Various model were used to\\ validate the existing \\ mathematical model.\end{tabular} &
  No use of AI model for prediction. \\ \hline
\begin{tabular}[c]{@{}c@{}}Kuniya et al.\\  {[}14{]}\end{tabular} &
  \begin{tabular}[c]{@{}c@{}}Estimation method is used for \\ prediction of transmission of\\  reproduction.\end{tabular} &
  \begin{tabular}[c]{@{}c@{}}Use to determine  only\\ basic reproduction of \\ Covid19 affection.\end{tabular} \\ \hline
\begin{tabular}[c]{@{}c@{}}Yuan et. al. \\ {[}15{]}\end{tabular} &
  \begin{tabular}[c]{@{}c@{}}In this work an internet search \\ based model was developed. \\ This model seemed to be useful\\  because there is a correlation \\ between the trends in daily new \\ cases and in new deaths and the \\ internet search terms.\end{tabular} &
  \begin{tabular}[c]{@{}c@{}}In spite of the correlation between \\ the search terms and trends of \\ death cases,  the accuracy was \\ very poor. Thus this \\ model should be further\\ developed.\end{tabular} \\ \hline
Ray et. al.{[}16{]} &
  \begin{tabular}[c]{@{}c@{}}In this work ensemble model\\ was introduced. Here multiple \\ model prediction was incorporated \\ into a combined forecast. \\ Ensemble model distill \\ information of multiple forecast.\end{tabular} &
  \begin{tabular}[c]{@{}c@{}}Ensemble forecast weighted\\ models equally which should \\ be done deeming the record of\\ performance. To remove the \\ uncertainty and raise the precision \\ longer horizon can be considered.\end{tabular} \\ \hline
\begin{tabular}[c]{@{}c@{}}Petropoulos \\ et. al.{[}17{]}\end{tabular} &
  \begin{tabular}[c]{@{}c@{}}In this exercise statistical forecast \\ was provided using robust time \\ series model.  The government \\ and individual can take steps\\ using the forecast.\end{tabular} &
  \begin{tabular}[c]{@{}c@{}}Here an assumption that future\\  will follow the past was made.\\  But this may not be true always.\end{tabular} \\ \hline
Pal et. al.{[}18{]} &
  \begin{tabular}[c]{@{}c@{}}This work introduced the use of \\ shallow LSTM.  This model \\ out-performs the state-of-the-art \\ models.\end{tabular} &
  \begin{tabular}[c]{@{}c@{}}Here weather data was combined\\ for prediction. Various modalities \\ of data can be combined like\\ flight, business, travellers etc.\end{tabular} \\ \hline
\begin{tabular}[c]{@{}c@{}}Rahman \\ et. al.{[}19{]}\end{tabular} &
  \begin{tabular}[c]{@{}c@{}}In this work, a large data set\\ of chest X-Ray was involved. \\ Image enhancement and \\ segmentationwere done which\\  were unprecedented. The standard \\ U-net was modified \\ which yielded a nice accuracy in \\ detection. This accuracy helped\\  detecting the spread of the virus.\end{tabular} &
  \begin{tabular}[c]{@{}c@{}}Though this work is a benchmark,\\ it arises complexity.\end{tabular} \\ \hline
\begin{tabular}[c]{@{}c@{}}Fouladi \\ et.al.{[}20{]}\end{tabular} &
  \begin{tabular}[c]{@{}c@{}}Classification of Covid-19 CT \\ images was performed using\\  modified VGG-16,  ResNet-50 etc.\\   DL models. The modifications of\\   parameters and hyper parameters\\  introduce  us with new models\\  for DL. The new models can be\\  utilized to classify Ebola, \\ SARS etc. images.\end{tabular} &
  \begin{tabular}[c]{@{}c@{}}Data augmentation was not\\ performed here. Larger data set \\ can be entailed. Quantity of\\ \\ infections should also be measured.\end{tabular} \\ \hline
\begin{tabular}[c]{@{}c@{}}Mahmoudi \\ et. el{[}21{]}\end{tabular} &
  \begin{tabular}[c]{@{}c@{}}The relationship between spread of \\ Covid-19 and population’s size \\ was studied.  This seems to be \\ necessary for policy makers.\end{tabular} &
  \begin{tabular}[c]{@{}c@{}}Fuzzy clustering was used\\ here. But regression model, \\ time series models and artificial\\ intelligence model could be \\ used here.\end{tabular} \\ \hline
\begin{tabular}[c]{@{}c@{}}Alok Negi \\ et. al.{[}27{]}\end{tabular} &
  \begin{tabular}[c]{@{}c@{}}Here using VGG16 and CNN on a \\ Simulated Masked Face Dataset\\ (SMFD), precautionary measures \\ are suggested.\end{tabular} &
  \begin{tabular}[c]{@{}c@{}}The model was not used for video\\ surveillance datasets.\end{tabular} \\ \hline
\begin{tabular}[c]{@{}c@{}}Alok Negi \\ et. al.{[}28{]}\end{tabular} &
  \begin{tabular}[c]{@{}c@{}}Using Keras-surgeon, model pruning\\ was done to simplify the neural \\ network. This model can be used to\\ detect mask wearing faces in public\\  places\end{tabular} &
  \begin{tabular}[c]{@{}c@{}}Comparison between Keras and\\  Pytorch was not presented.\end{tabular} \\ \hline
\begin{tabular}[c]{@{}c@{}}Alok Negi \\ et. al. {[}29{]}\end{tabular} &
  \begin{tabular}[c]{@{}c@{}}To detect malaria infection CNN \\ model was developed. Result was \\ very satisfactory.\end{tabular} &
  \begin{tabular}[c]{@{}c@{}}Comparison between Keras and \\ Pytorch was not present.\end{tabular} \\ \hline
\begin{tabular}[c]{@{}c@{}}Krishnan \\ Kumar et. al.\\ {[}30{]}\end{tabular} &
  \begin{tabular}[c]{@{}c@{}}Facilitates the summarization of \\ multi-view video.\end{tabular} &
  \begin{tabular}[c]{@{}c@{}}Mono-view video summarization \\ is not possible here.\end{tabular} \\ \hline
\begin{tabular}[c]{@{}c@{}}Haroon\\ Ansari et. al.\\ {[}31{]}\end{tabular} &
  \begin{tabular}[c]{@{}c@{}}Using Hidden Markov Model the \\ depression of human mind is \\ detected.\end{tabular} &
  \begin{tabular}[c]{@{}c@{}}The procedure may yield result \\ as depression in place of \\ exhilaration.\end{tabular} \\ \hline
\begin{tabular}[c]{@{}c@{}}Anshuman \\ Darbari et. al.\\ {[}32{]}\end{tabular} &
  \begin{tabular}[c]{@{}c@{}}Encourages the use of AI in the \\ field of thoracic surgery.\end{tabular} &
  \begin{tabular}[c]{@{}c@{}}Practical thoracic surgery may not \\ not be feasible by AI\end{tabular} \\ \hline
\begin{tabular}[c]{@{}c@{}}Alok Negi\\ et. al.{[}33{]}\end{tabular} &
  \begin{tabular}[c]{@{}c@{}}Various plant diseases were detected\\ using CNN\end{tabular} &
  \begin{tabular}[c]{@{}c@{}}This deep learning model can be\\ used for only one type of plant \\ classification.\end{tabular} \\ \hline
\begin{tabular}[c]{@{}c@{}}Alok Negi\\ et. al. {[}34{]}\end{tabular} &
  \begin{tabular}[c]{@{}c@{}}The diseases of citrus fruit and \\ leaves were detected using CNN.\\ This was a good job to detect the \\ defected plants and take proper\\ medical treatment.\end{tabular} &
  \begin{tabular}[c]{@{}c@{}}This defect detection can be \\ \\ compared with other deep learning \\ models.\end{tabular} \\ \hline
\begin{tabular}[c]{@{}c@{}}Qingzhen Xu.\\ {[}39{]}\end{tabular} &
  \begin{tabular}[c]{@{}c@{}}A machine learning method is \\ improved by a two dimensional\\ equation is found by finite difference\\ method. This method can be used \\ for financial prediction.\end{tabular} &
  \begin{tabular}[c]{@{}c@{}}This technique was not applied in \\ the field of science and technology.\end{tabular} \\ \hline
\begin{tabular}[c]{@{}c@{}}Qingzhen Xu\\ et.al.{[}40{]}\end{tabular} &
  \begin{tabular}[c]{@{}c@{}}For analyzing the complex images,\\ saliency detection model is produced\\ through a new method.\end{tabular} &
  \begin{tabular}[c]{@{}c@{}}Comparison among various \\ methods is not presented.\end{tabular} \\ \hline
\begin{tabular}[c]{@{}c@{}}Qingzhen Xu\\ et. al.{[}41{]}\end{tabular} &
  \begin{tabular}[c]{@{}c@{}}Fall of osteoporosis is predicted by \\ transfer learning with s high \\ accuracy.\end{tabular} &
  \begin{tabular}[c]{@{}c@{}}Comparison among various\\ methods is not presented.\end{tabular} \\ \hline
\begin{tabular}[c]{@{}c@{}}Qingzhen Xu\\ et.al.{[}42{]}\end{tabular} &
  \begin{tabular}[c]{@{}c@{}}Prediction of trend of stock based on\\ money flow using fuzzy clustering\\ and collaborative filtering can be \\ possible from this work\end{tabular} &
  \begin{tabular}[c]{@{}c@{}}Considering the investor\\ characteristics this model can be\\ improved.\end{tabular} \\ \hline
\end{longtable}
\end{landscape}

Below table shows the comparison of the proposed method compare to others. 

\begin{table}[h!]
\centering
\caption{Comparison among Deep Learning Models}
\label{tab:my-table1}
\resizebox{\textwidth}{!}{%
\begin{tabular}{|l|l|l|}
\hline
Model & Advantage & Limitation \\ \hline
LSTM &
  \begin{tabular}[c]{@{}l@{}}1. Capable of learning long-range dependencies.\\ 2. Able to solve dependencies problem.\\ 3. Work as a memory network with large dataset.\end{tabular} &
  \begin{tabular}[c]{@{}l@{}}1. Complex architecture.\\ 2. Repeating module has complex layers.\\ 3. Unable to remove vanishing gradients issue completely\end{tabular} \\ \hline
GRU &
  \begin{tabular}[c]{@{}l@{}}1. Use less memory compared to LSTM approach\\ 2. Faster training and execution capability.\end{tabular} &
  \begin{tabular}[c]{@{}l@{}}1. Able to train less parameters for training purpose.\\ 2. Provides lower accuracy.\end{tabular} \\ \hline
Linear Regression &
  \begin{tabular}[c]{@{}l@{}}1. Provides better result when data sets are linear.\\ 2.  Using dimensional reduction over-fitting can be avoided.\end{tabular} &
  \begin{tabular}[c]{@{}l@{}}1. Under-fitting could not be mitigated.\\ 2. Lower accuracy with poor performance.\end{tabular} \\ \hline
\end{tabular}%
}
\end{table}

\section{Proposed Methodology}

In this section, the methodology of the research proposed are divided into three parts. The data collection for this work is describe in section \ref{datacollection}. Section \ref{minmaxsclar} describe the MinMax Scalar method to format the collected data. Then prepare the data for LSTM deep learning model for training and testing purpose. The details of LSTM architecture is given in section \ref{LSTMarchitecture}. \\
At first the data is read as data frame from CSV file. The graph of cumulative death per day and change of death per day are plotted. The data is partitioned as training and testing data sets. A predictive model is constructed by LSTM. Using this model and training data, anticipation is accomplished. The prediction and testing data are compared and checked for validation of the proposed model. Later, the whole data is used in the model. Again the validation step is performed. Finally, for comparative exhibition of historical and predictive daily cases, a figure is plotted.

\subsection{Data Collection and Description}
\label{datacollection}

An updated recovery cases data set were  collected from Github repository which automatically update the Covid-19 data in daily cases. The time series data have sorted  for prediction purpose. The data of the repository handled by the CSSE(Central for System Science and Engineering) of Johns Hopkins University.  The data-sets contain the recovery cases across global as a csv data file between 22 January, 2020 to 27 February, 2021 \cite{c18}. The data-set contains 258 regions recovery cases per day due to Covid-19 pandemic. 

\begin{figure}[htbp]
    \centering
    \includegraphics[width=1\textwidth,height=8cm]{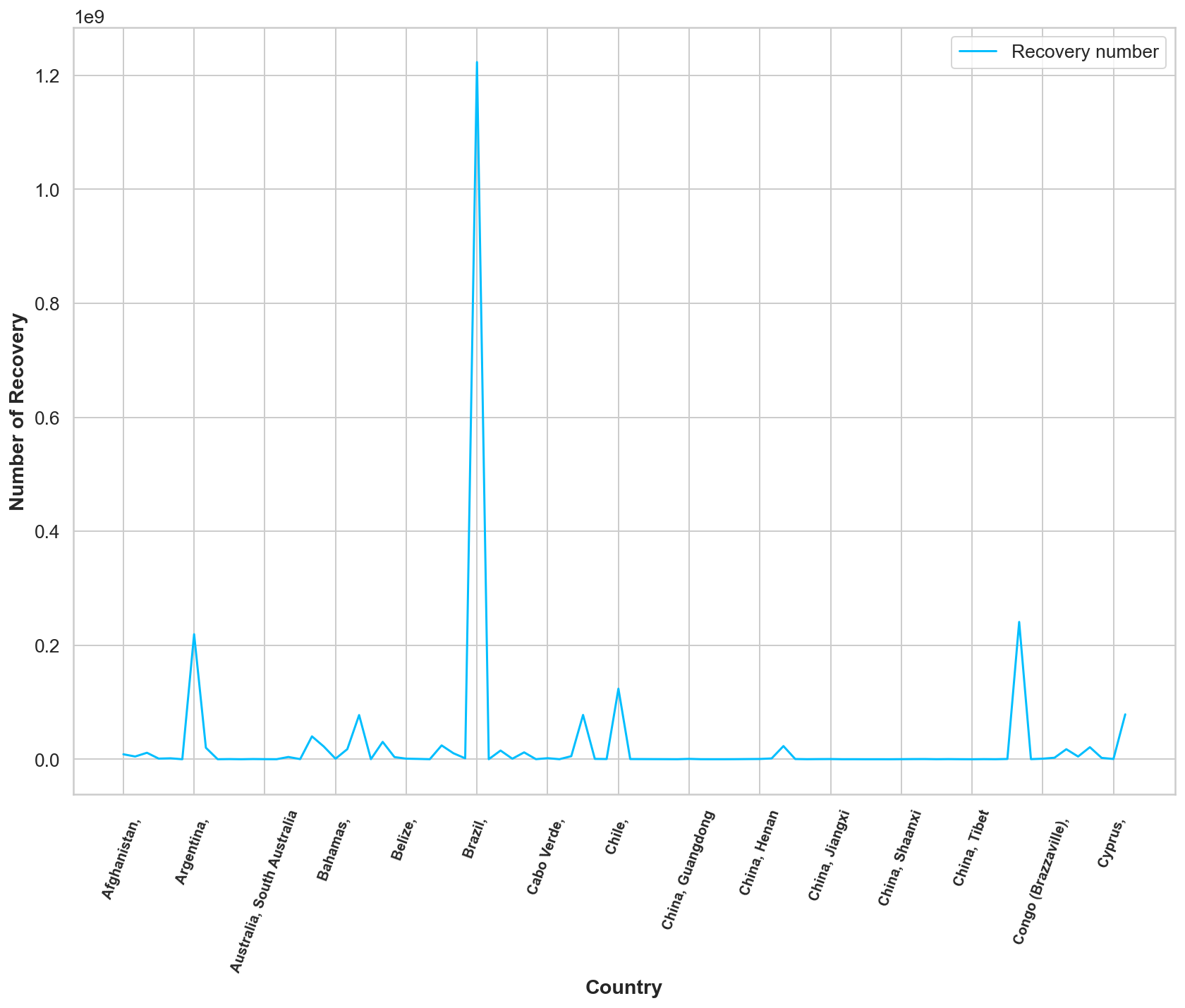}
    \caption{Recovery cases for various countries. The countries are alphabetically arranged from Afghanistan to Cyprus}
    \label{f2}
\end{figure}
\begin{figure}[htbp]
    \centering
    \includegraphics[width=\textwidth,height=7cm]{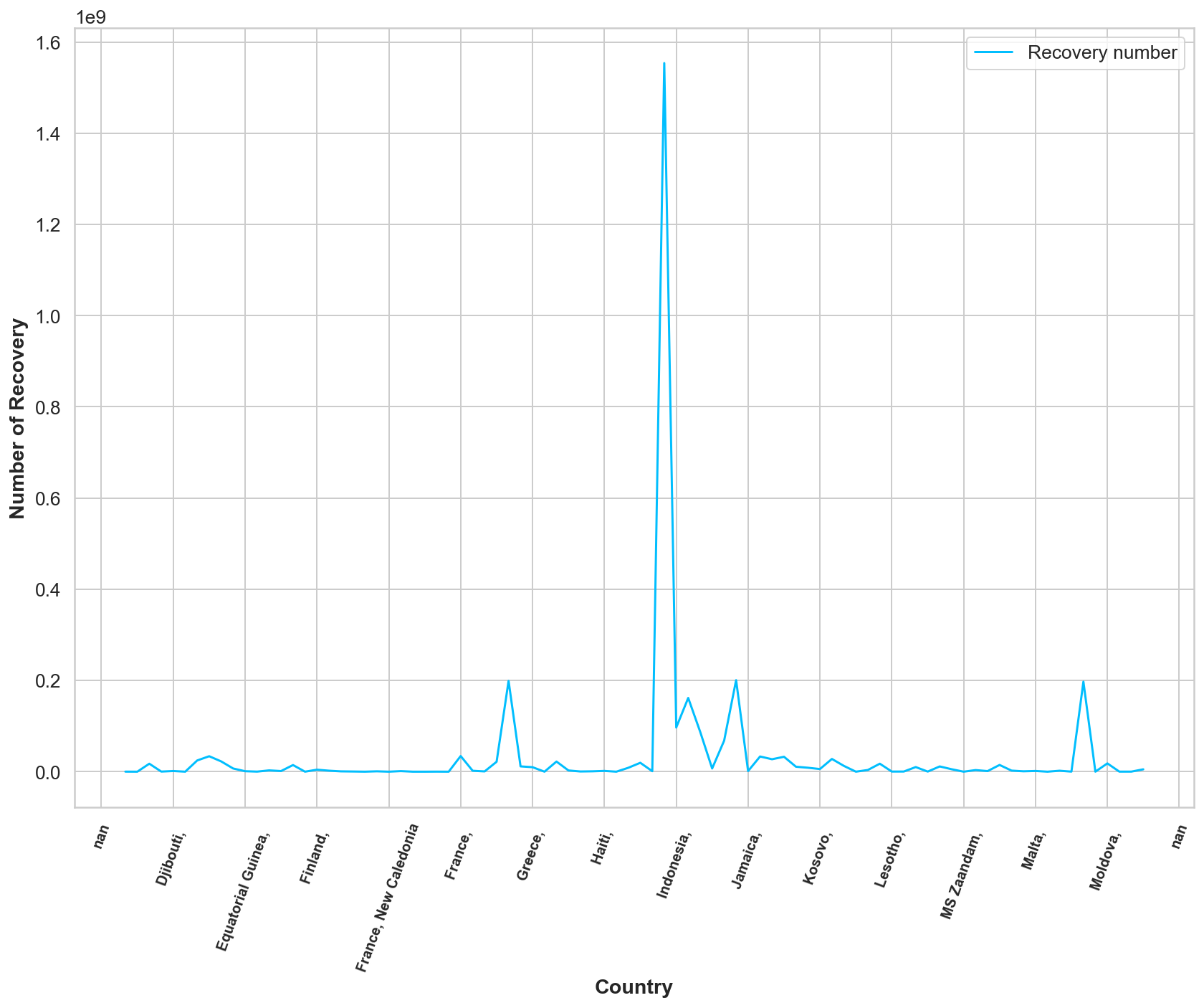}
    \caption{Recovery cases for various countries. The countries are alphabetically arranged from Djibouti to Moldova}
    \label{f3}
\end{figure}
\begin{figure}[htbp]
    \centering
    \includegraphics[width=1\textwidth,height=7cm]{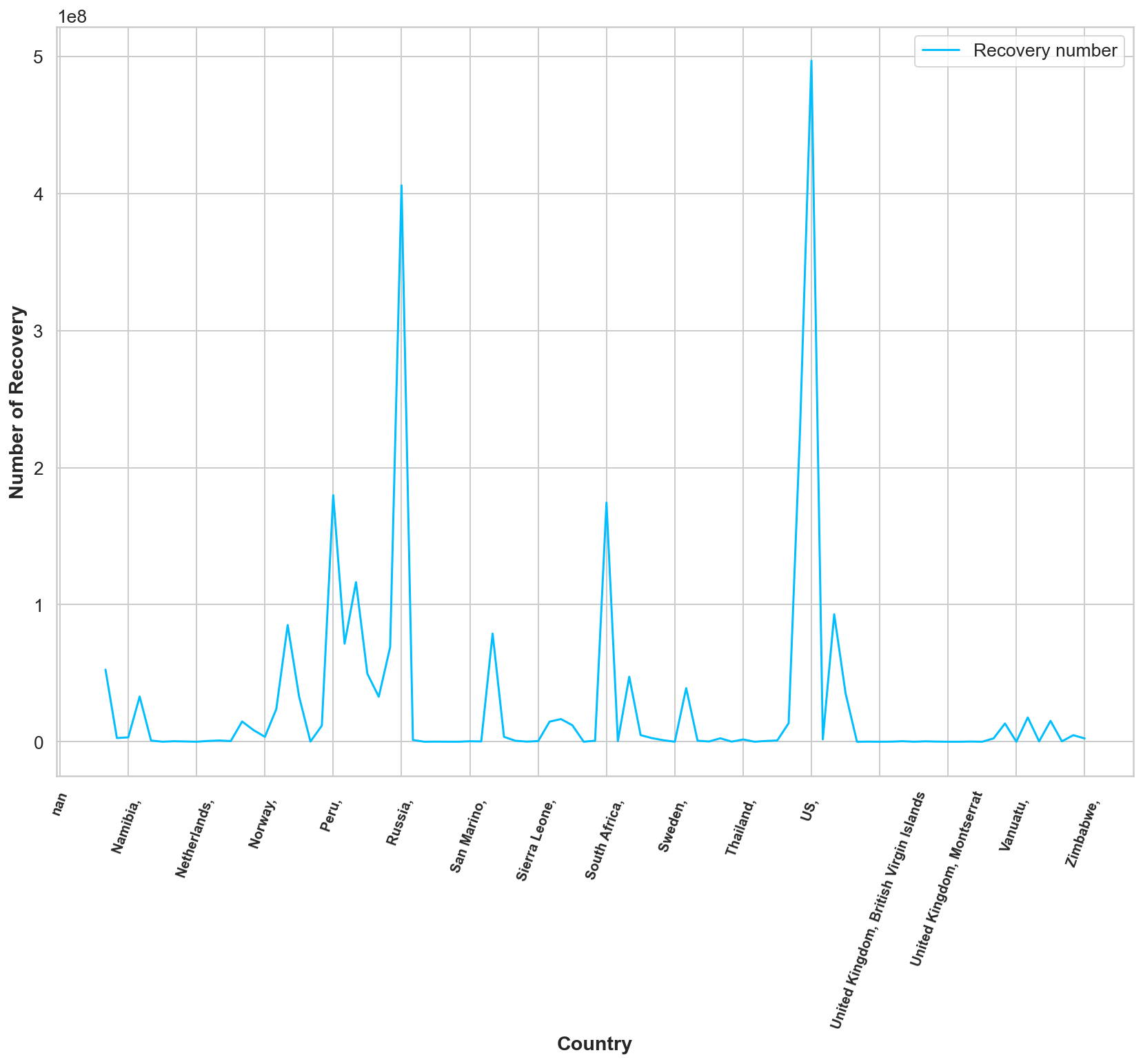}
    \caption{Recovery cases for various countries. The countries are alphabetically arranged from Namibia to Zimbabwe}
    \label{f4}
\end{figure}

Figure \ref{f2},\ref{f3} ,\ref{f4} shows the overview of frequent recovery happens in countries using Raw data. This picture indicates that India is the highest recovering country wheres Brazil is second in recovering of life. Other countries such as U.S., Russia exhibit moderate amount of recovery. In every moment people are migrating from each country to others. Therefore, the countries should take proper steps to stop the infected people migration from neighbouring countries. Otherwise, the nearest country will have to pay high death toll. 

\subsection{Feature Scaling}
\label{minmaxsclar}

The machine learning algorithm do not perform well with different scale data. To avoid this issue, the min-max  features scaling transform perform well to scale the data in the same platform \cite{c19}. Sometime, its known as normalization since its normalize the between 0 to 1. Its operation can be expressed using the following procedure:
\begin{equation}\label{masud}
 X_{sc}=\frac{X-X_{min}}{X_{max}-X_{min}}. 
\end{equation}

Where $X$ is the original data-set. $X_{min}$ is the minimum value in the data range whereas $X_{max}$ determine the maximum value of the data. 

\subsection{Architecture of Long Short Term Memory (LSTM)}
\label{LSTMarchitecture}

\begin{figure}[htbp]
    \centering
    \includegraphics{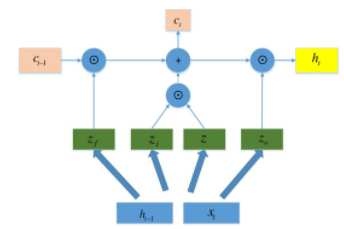}
    \caption{General Architecture of LSTM}
    \label{f5}
\end{figure}

LSTM (Long Short Term Memory) is mainly approached to predict the future tread trend of confirm death case using history of the tanning data. LSTM is the one kind of RNN network. Generally, RNN works on short-term memory basis wheres LSTM on short and long-term memory. The major drawback of RNN is the disappearance of gradients descent. LSTM address this issue by combining the long-term memory with short-term through the subtle control gate. In 1997, LSTM was proposed by Hochreiter and Schmidhuber \cite{c20}. Figure \ref{f5}  shows the general architecture of the LSTM model. Three gates are basically used named as input gate, forget gate and output gate. Initially, the input data is feed to the forget through input gate. The forget gate decide which information is discard by reading the hidden state, $h_{t-1}$ and input layer $x_t$ information.It generates an output vector between zero and 1. This value of vector determine how much amount of information is discarded or retained in the state cell $C_t$.The zero  value of the vector indicates all  information is discarded and one means it retain all information. 
After that, there is two steps to identify how much information is added to the cell. First, through an input gate operation $h_{t-1}$, and  $x_{t}$ decide which information to update.
\begin{equation}\label{eqexpmuts}
z_{f}=\sigma(W_{f}.[h_{t-1}, x_{t}]+b_{f}).
\end{equation}
\begin{equation}
z_{i}=\sigma(W_{i}.[h_{t-1}, x_{t}]+b_{i}).
\end{equation}
The new candidate cell $Z$ is acquired through a $tanh$ layer using $h_{t-1}$, and  $x_{t}$. 
\begin{equation}\label{eqexpmuts1}
z=tanh(W.[h_{t-1}, x_{t}]+b).
\end{equation}
In the later stage, the cell is updated as follows:
\begin{equation}\label{eqexpmuts2}
    c_{t}=z_{f}*C_{t-1}+z_{i}*z.
\end{equation}
Where * and + represent vector multiplication and addition.
At last, the output of the cell should be identified after updating. This is done via the help of $h_{t-1}$, and  $x_{t}$. An Sigmoid activation function output gate is applied to mapping the output. Than, to obtain a vector between +1 to -1, the cell state pass through the $tanh$ layer.    
\begin{equation}
\label{eqexpmuts3}
    z_{0}=\sigma(W_{o}.[h_{t-1}, x_{t}]+b_{0}).
\end{equation}
\begin{equation}
    h_{t}=z_{o}*tanh(c_{t}). 
\end{equation}
where $z_{f}$,  $z_{o}$ and $z_{i}$ are the  gate control state of the forget gate, output gate, and input gate, respectively. Also, z is known as candidate input cell  through the layer of $tanh$. 
\begin{equation}\label{eqexpmuts4}
z_{f}=\sigma(W_{f}.[h_{t-1}, x_{t}]+b_{f}). 
\end{equation}
\begin{equation}
    z_{i}=\sigma(W_{i}.[h_{t-1}, x_{t}]+b_{i}).
\end{equation}

\section{Results and Discussion}

\begin{figure}[htbp]
    \centering
    \includegraphics[width=1\textwidth, height=5cm]{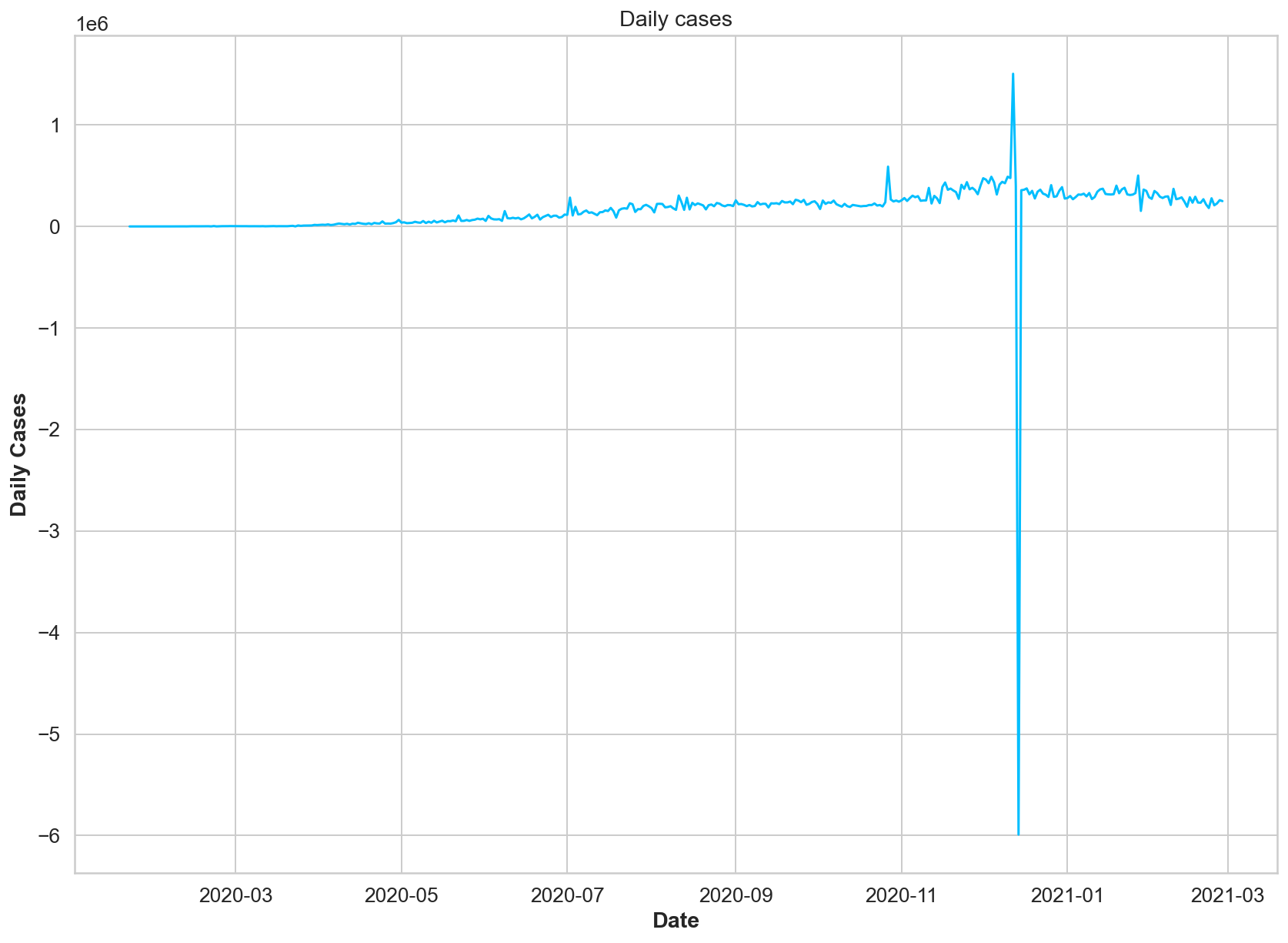}
    \caption{Daily recovery cases across global according to real data.}
    \label{f6}
\end{figure}

\begin{figure}[htbp]
    \centering
    \includegraphics[width=1\textwidth, height=5cm]{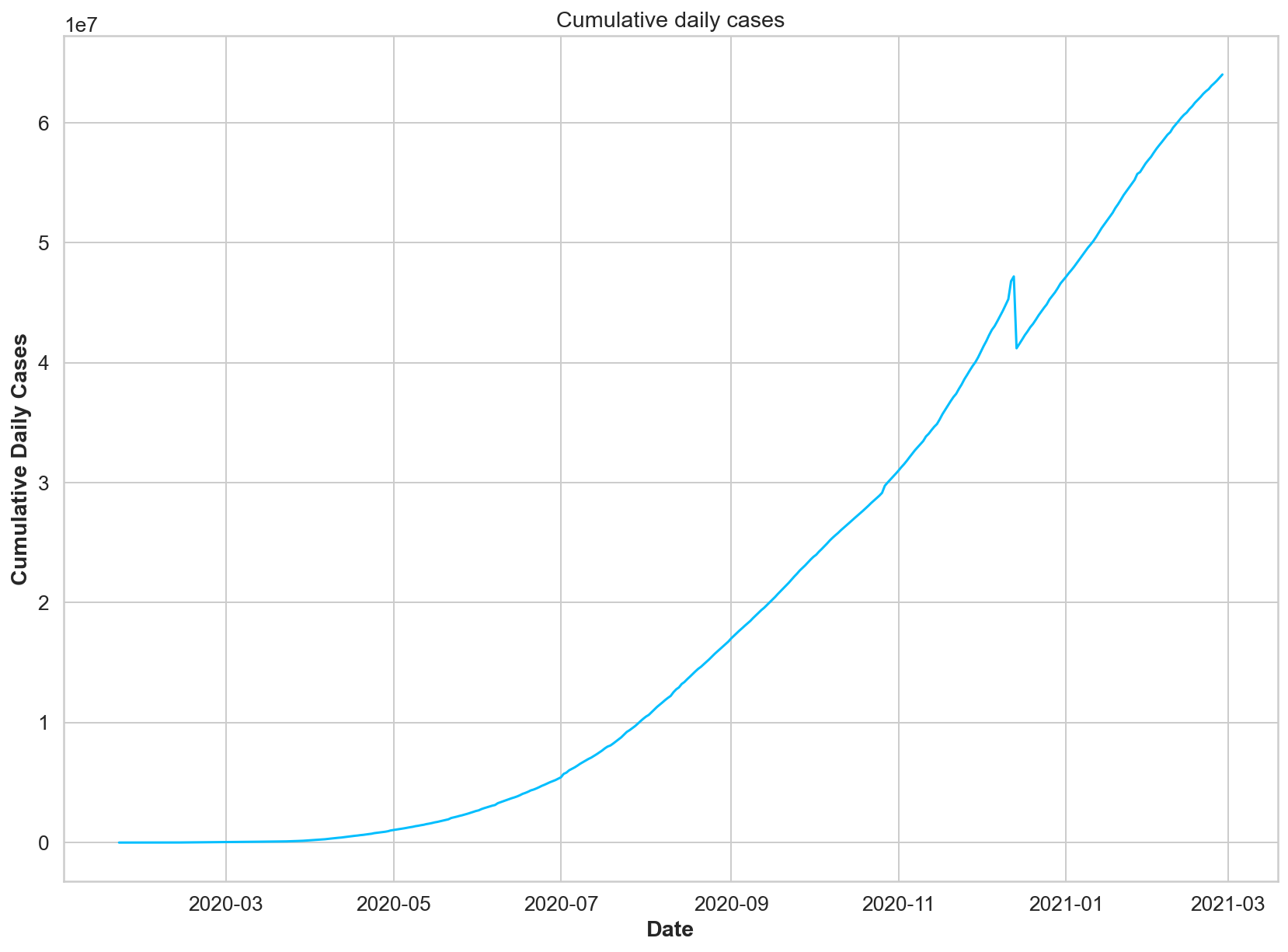}
    \caption{Cumulative case of daily recovery across globe.}
    \label{f7}
\end{figure}

\begin{figure}[htbp]
    \centering
    \includegraphics[width=1\textwidth, height=5cm]{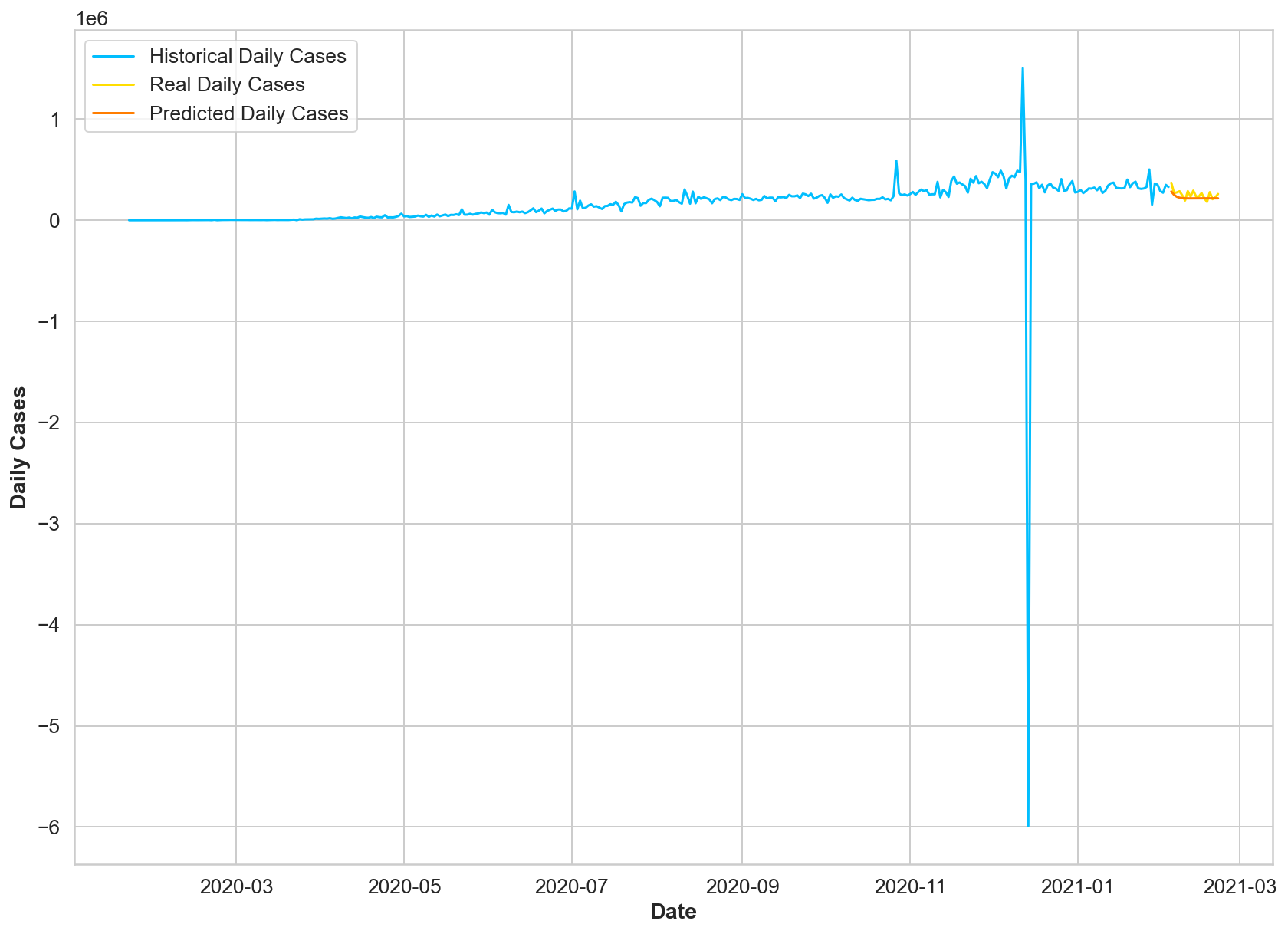}
    \caption{Combined show of Historical, Real and Predicted daily recovery cases.}
    \label{f8}
\end{figure}

\begin{figure}[htbp]
    \centering
    \includegraphics[width=1\textwidth, height=5cm]{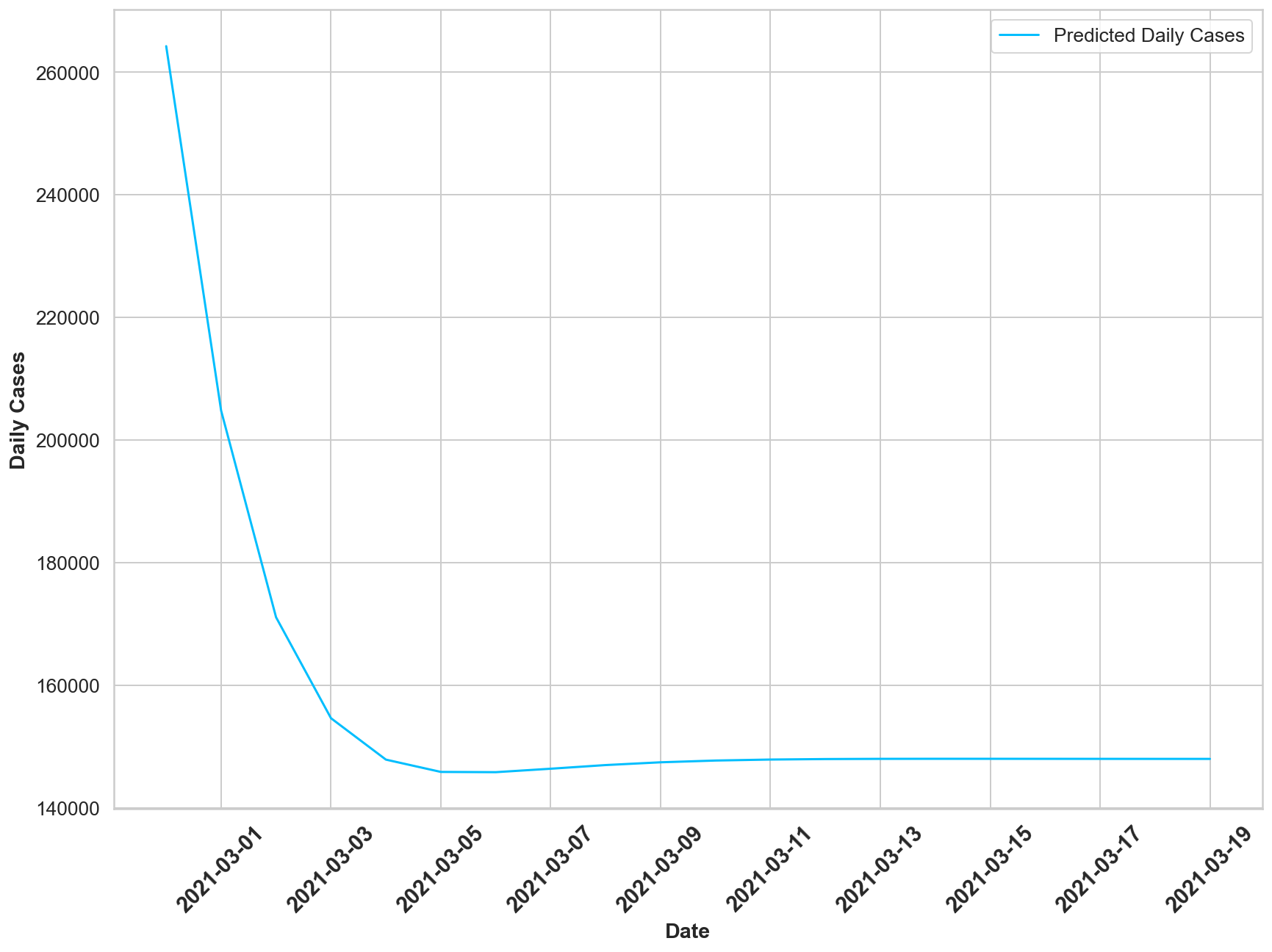}
    \caption{Predicted case using all training data for 20 days}
    \label{f9}
\end{figure}

This section presents all computational results related to this research work.\\ 
Figure \ref{f6} shows the graphical visualization of changing in the number of recovery cases for 403 days. In this figure, the cumulative recovery case of 22 January, 2020 used as a reference. This value is deduced from the cumulative deaths of other days. Thus from this figure we find that at some days the number of death increases and at other days the number decreases. 
The cumulative case of everyday recovery case is figured out in Figure~\ref{f7}. In this case, the numbers of recovery per day from 22 January 2020 to 01 January, 2021 of 271 regions in the world are plotted. From this figure, it is concluded that, the number of recovery is increased most of the days.
The comparative study of the historical, real and predicted results are presented in Figure \ref{f8}. This figure tells that the historical daily cases obtained from 379 days of training data and the rest of 24 days data used as a real case. Based on historical data, the prediction is done using LSTM approach. For the 24 days the number of predicted recoveries decreases slightly compared to real cases. 
Figure \ref{f9} shows the prediction result of daily cases using whole data sets. The prediction is for 20 days with 60 number of epochs and 24 batch size. In this cases, the predicted daily cases decaying fast until 03 March 2021 and than obtain an almost stable position till 19 March 2021.

\begin{figure}[htbp]
    \centering
    \includegraphics[width=1\textwidth, height=5cm]{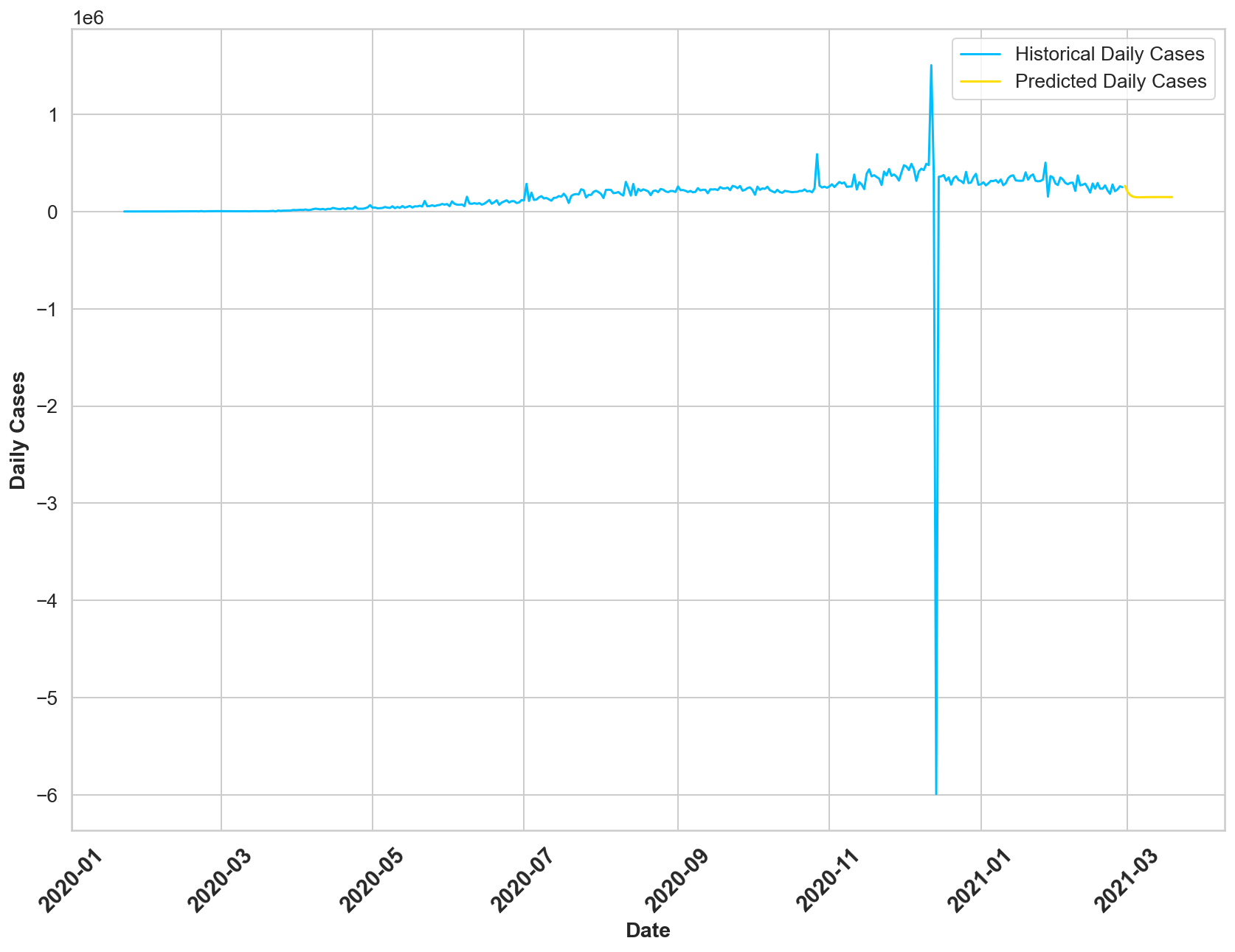}
    \caption{Combined show of historical and predicted daily recovery cases.}
    \label{f10}
\end{figure}

The historical and predicted daily cases is shown Figure \ref{f10} with considering whole data sets. The number of prediction is decrease. This may be happened due to majority of days provides falling recovery history. 

\section{Conclusions}
There is a contingency of spark of this global issue. Still many people are neglecting the panic. So we must need the proper decision. Forecasting and uncertainty are the integral part of decision making. Tentative anticipation of future is not only a concern of health sector but also of politics and economy. Our study is beneficial for this prediction. Our one can be used to combat against the second wave. This algorithm can be more affluent if more update data is available. Lacking of our anticipation emerges from demography, hospital capacity and social distances.









\section*{Declarations}

The following sections contain the information regarding the declaration.
\begin{itemize}
\item{Funding:} Not Applicable
\item{Conflict of interest/Competing interests:}The authors declare that there is no conflict of interest.
\item{Ethics approval:}This work has not been published anywhere yet.
\item{Consent to participate:}Everyone among us opined to participate in this study.
\item{Consent for publication:}Everyone of us is agreed to publish the article in this journal.
\item{Availability of data and materials:}Not Applicable.
\item{Code availability:}The code is not available in the cloud.
\item{Authors' contributions:}All the authors have equal contribution in this article.
\end{itemize}

\bibliographystyle{unsrt}  

\end{document}